\documentclass[10pt,twocolumn,letterpaper]{article}

\usepackage{iccv}
\usepackage{times}
\usepackage{epsfig}
\usepackage{graphicx}
\usepackage{amsmath}
\usepackage{amssymb}
\usepackage{bm}
\usepackage{subcaption}
\usepackage{multirow}
\usepackage{pifont}

\usepackage[breaklinks=true,bookmarks=false]{hyperref}

\iccvfinalcopy 

\def\iccvPaperID{6496} 

\ificcvfinal\fi
\begin{document}

\title{Weakly-supervised Action Localization with Background Modeling}

\author{Phuc Xuan Nguyen\\
University of California, Irvine\\
{\tt\small nguyenpx@ics.uci.edu}
\and
Deva Ramanan\\
Carnegie Mellon University\\
{\tt\small deva@cs.cmu.edu}
\and
Charless C. Fowlkes\\
University of California, Irvine\\
{\tt\small fowlkes@ics.uci.edu}
}

\maketitle

\begin{abstract}
We describe a latent approach that learns to detect actions in long sequences
given training videos with only whole-video class labels. Our approach makes
use of two innovations to attention-modeling in weakly-supervised learning.
First, and most notably, our framework uses an attention model to extract both
foreground and background frames whose appearance is explicitly modeled. Most
prior works ignore the background, but we show that modeling it allows our
system to learn a richer notion of actions and their temporal extents. Second,
we combine bottom-up, class-agnostic attention modules with top-down,
class-specific activation maps, using the latter as form of self-supervision
for the former. Doing so allows our model to learn a more accurate model of
attention without explicit temporal supervision. These modifications lead to
$~10\%$ $AP@IoU$=0.5 improvement over existing systems on THUMOS14. Our
proposed weakly-supervised system outperforms recent state-of-the-arts by at
least $4.3\%$ $AP@IoU$=0.5. Finally, we demonstrate that weakly-supervised
learning can be used to aggressively scale-up learning to in-the-wild,
uncurated Instagram videos. The addition of these videos significantly improves
localization performance of our weakly-supervised model.
\end{abstract}

\section{Introduction}

We explore the problem of weakly-supervised action localization, where the task
is learning to detect and localize actions in long sequences given videos with
only video-level class labels. Such a formulation of action understanding is
attractive because it is well-known that precisely estimating the start and end
frames of actions is challenging even for humans~\cite{CabaJHG18}. We build on
a body of work that makes use of attentional processing to infer frames most
likely to belong to an action. We specifically introduce the following
innovations.
\begin{figure}[t]
\captionsetup{font=small}
\centering
\includegraphics[width=1\linewidth]{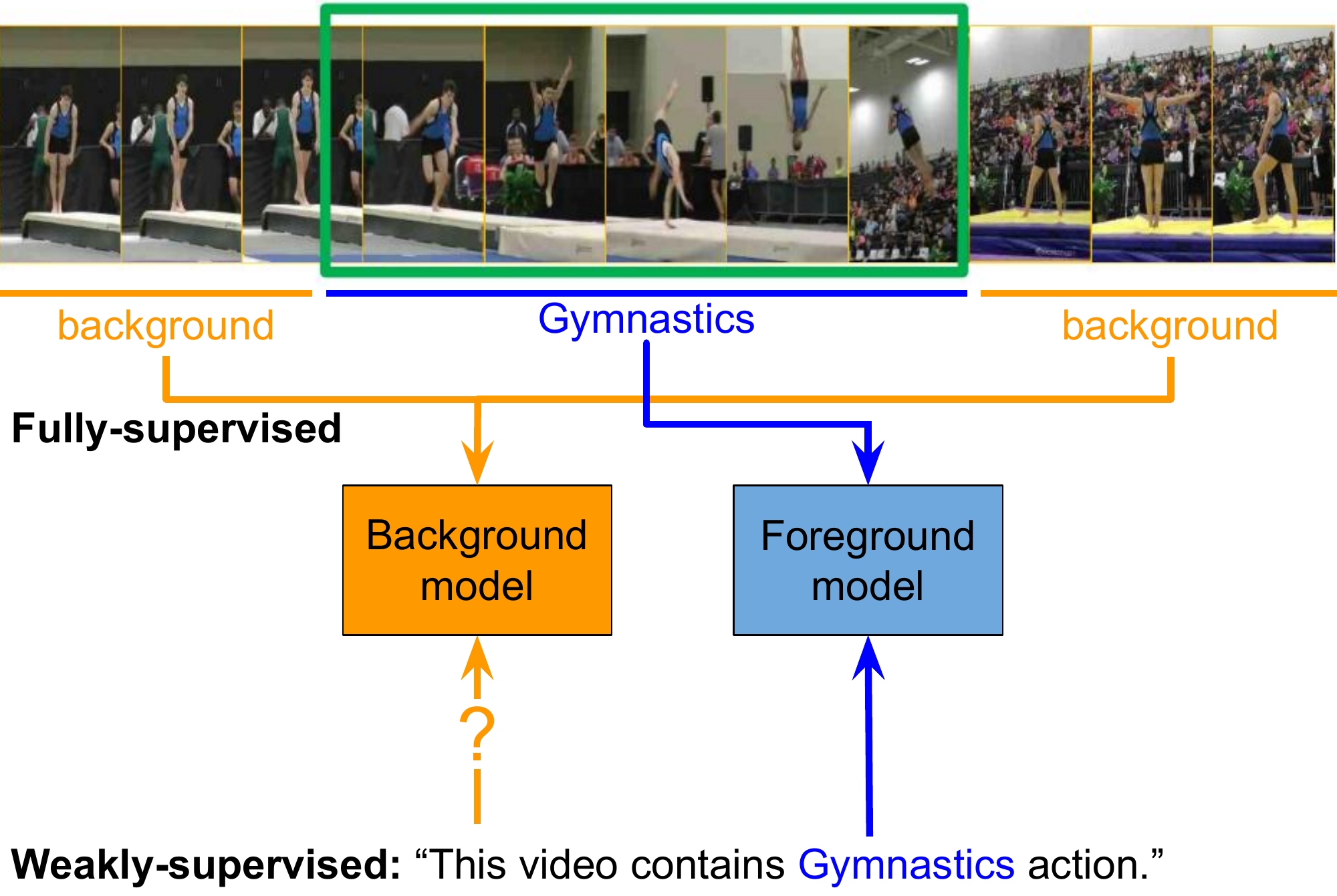}
\caption{With fully-supervised data where exact boundaries of actions are
	provided, we can train highly discriminative detection models that use
	background regions as negative examples, implicitly modeling background
	content. In weakly-supervised setting where only video-level labels are
	known, current approaches simply train a foreground model to respond
	strongly at some locations within the video, but leave the remaining
	background frames unmodeled. In this paper we show that a model which
	explicitly accounts for background frames substantially improves on
	weakly-supervised localization.}
\label{fig:overview}
\end{figure}

{\bf Background modeling:} Classic pipelines use attentional pooling to focus a
model on those frames likely to contain the action of interest. We show that by
modeling the remaining background frames, one can significantly improve the
accuracy of such methods. Interestingly, fully-supervised systems for both
objects~\cite{ren2015faster} and actions~\cite{chao2018rethinking} tend to
build explicit models (or classifiers) for background patches and background
frames, but this type of reasoning is absent in most weakly-supervised systems.
Notable exceptions in the literature include probabilistic latent-variable
models that build generative models of both foreground and
background~\cite{jojic2001learning}. We incorporate background modeling into
discriminative network architectures as follows: many such networks explicitly
compute an attention variable, $\lambda_{t}$, that specifies how much frame $t$
should influence the final video-level representation (by say, weighted pooling
across all frames). Simply put, we construct a pooled video-level feature
that focuses on the background by weighing frames with $1- \lambda_t$.

{\bf Top-down guided attention:} Our second innovation is the integration of
top-down attentional cues as additional forms of supervision for learning
bottom-up attention. The attention variable $\lambda_t$, typically
class-agnostic, looks for generic cues that apply to all types of actions. As
such, it can be thought of as a form of bottom-up attentional
saliency~\cite{gao2008discriminant}. Recent works have shown that one can also
extract {\em top-down} attentional cues from classifiers that operate on pooled
features by looking at (temporal) class activation maps (T-CAM)
~\cite{nguyen2017weakly,zhou16learning}. We propose to use class-specific
attention maps as a form of supervision to refine the bottom-up attention maps
$\lambda_t$. Specifically, our loss encourages bottom-up attention maps to agree with
top-down class-specific attention map (for classes {\em known} to exist in a
given training video). 

{\bf Micro-videos as training supplements:} We observe there is a
huge influx of microvideos on social media platforms (Instagram,
Snapchat)~\cite{nguyen2016open}. These videos often come with user-generated
tags, which can be loosely viewed as video-level labels. This type of data
appears to be an ideal source for weakly-supervised video training data.
However, the utility of these videos remains to be established. In this paper, we show
that the addition of microvideos to existing training data allows aggressive
scaling up of learning which improves action localization accuracy.

Our contributions are summarized below:
\begin{itemize}
\item We extend prior weakly-supervised action localization systems to include
  background modeling and top-down class-guided attention.
\item We present extensive comparative analyses between our models versus other
  state-of-the-art action localization systems, both weakly-supervised and
    fully-supervised, on THUMOS14~\cite{jiang14thumos} and
    ActivityNet~\cite{activitynet}.
\item We demonstrate the promising effects of using microvideos as supplemental,
  weakly-supervised training data.
\end{itemize}

\section{Related Works}

In recent years, progress in temporal action localization has been driven by
large-scale datasets such as THUMOS14~\cite{jiang14thumos},
Charades~\cite{sigurdsson2016hollywood}, ActivityNet~\cite{activitynet} and
AVA~\cite{gu2017ava}. Building such datasets has required substantial human
effort to annotate the start and end points of interesting actions
within longer video sequences. Many approaches to fully-supervised action
localization leverage these annotations and adopt a two-stage,
propose-then-classification framework~\cite{buch2017sst, shou16temporal,
escorcia2016daps, heilbron16fast, shou17cdc, zhao2017temporal}.  More recent
state-of-the-art methods~\cite{gao2017turn, gao2017cascaded, xu17r,
dai2017temporal, chao2018rethinking} borrow intuitions from the recent object
detection frameworks (e.g. R-CNN). One common factor among these approaches
is using non-action frames within the video for building background model.

Temporal boundary annotations, however, are expensive to obtain. This motivates
efforts in developing models that can be trained with weaker forms of
supervision such as video-level labels.
UntrimmedNets~\cite{wang17untrimmednets} uses a classification module to perform
action classification and selection module to detect important temporal
segments. Hide-n-Seek~\cite{singh17hide} addresses the tendency of popular
weakly-supervised solutions - networks with global average pooling - to only
focus on the most discriminative frames by randomly hiding parts of
the videos.  STPN~\cite{nguyen2017weakly} introduced an attention module to
learn the weights for the weighted temporal pooling of segment-level feature
representations. This method generates detections by thresholding Temporal Class
Activation Mappings (T-CAM) weighted by the attention values.
AutoLoc~\cite{shou2018autoloc} introduces a boundary predictor to predict
segment boundaries using an anchoring system. The boundary predictor is driven
by the Outer-Inner-Constrastive Loss, which encourages segments with high
activation on the inside and weaker activations on the immediate neighborhood
of this segment. W-TALC~\cite{paul2018w} introduces a system with k-max
Multiple Instance Learning and explicitly identifies the correlations between
videos of similar categories by a co-activity similar loss. None of the
aforementioned methods attempts to explicitly model background content during
training.

\begin{figure}
\captionsetup{font=small}
\centering
\includegraphics[width=1\linewidth]{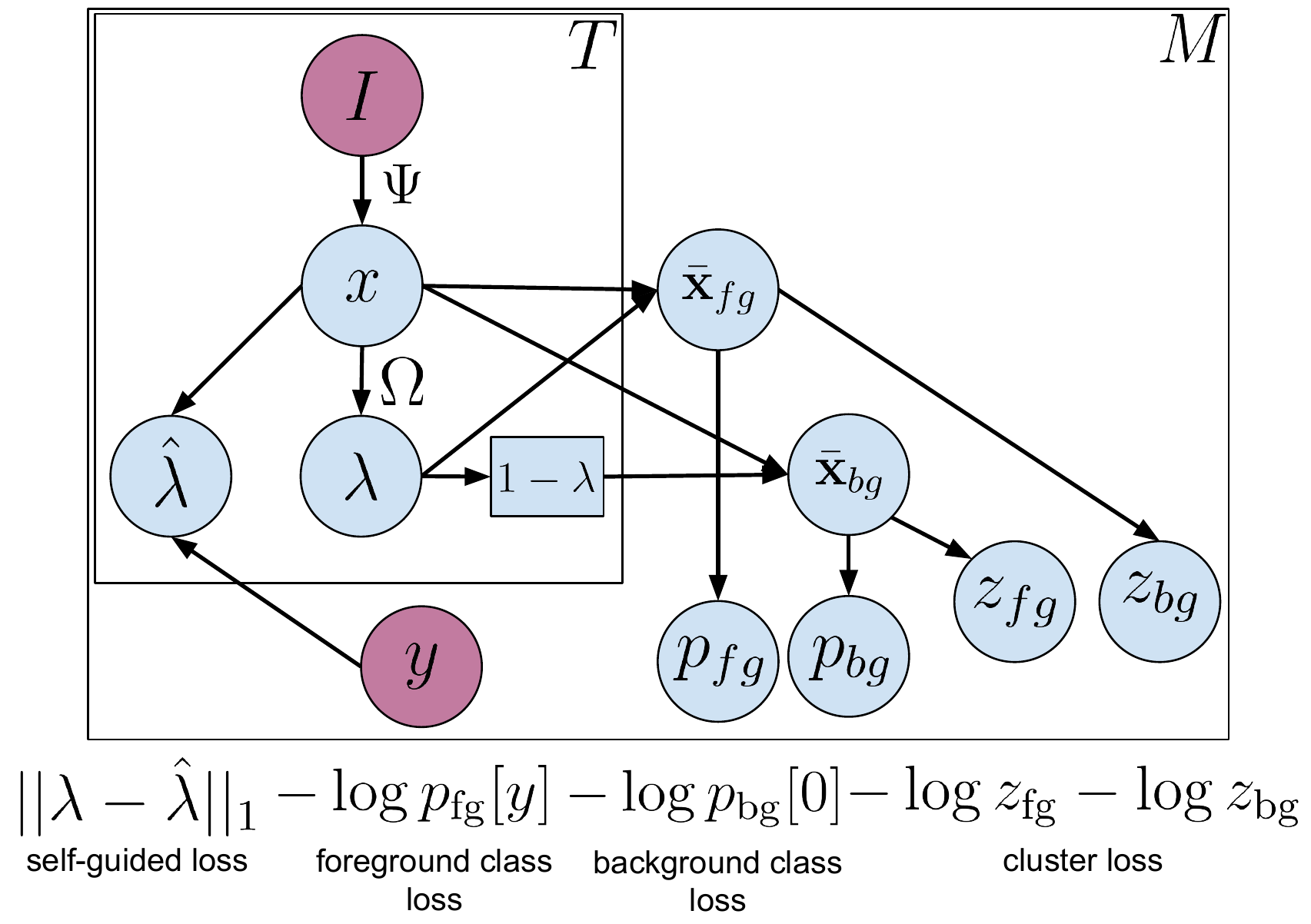}\\
	\caption{Network architecture for our weakly supervised action
	localization model. Using a pre-trained network, we extract the
	features representation for short video segments. The attention module
	$\Omega$ predicts frame level attention $\lambda$ which can be used to
	pool the frame-level features into a single foreground video-level
	feature representation. The complement of the attention vector,
	$1-\lambda$, can also be used to pool segments belonging to the
	background into a video-level background representation. Video-level
	labels are predicted from these pooled features. In addition to this
	action-specific top-down model appearance, we also include bottom-up
	clustering loss which asserts that the video should segment into
	distinct foreground and background appearances $z_{fg},z_{bg}$. To link
	these two, we compute an attention target ${\hat \lambda}$ based on the
	class activations of the ground-truth video label $y$ using a
	``self-guided'' loss that encourages the predicted attention $\lambda$
	to match this target.}
\label{fig:architecture}
\end{figure}

\section{Localization from Weak Supervision}

Assume we are provided with a training set of videos and video-level labels $y
\in \{0,\ldots,C\}$, where $C$ denotes the number of possible actions and $0$
indicates no action (the background). In each frame $t$ of each video, let us
write ${\bf x}_t \in \mathbb{R}^d$ for a feature vector based on RGB and
optical flow extracted at that frame (e.g., pretrained on a related video
classification task). We then can write each training video as a tuple of
feature vectors and video-level label: $$( \{ {\bf x}_t \}, y), \quad {\bf x}_t
\in \mathbb{R}^d, y \in \{0,\ldots,C\}$$

In principle, videos may contain multiple types of actions, in which case it is
more natural to model $y$ as a multi-label vector. From this set of {\em
video-level} training annotations, our goal is to learn a {\em frame-level}
classifier that can identify which of the $C+1$ actions (or background) is
taking place at each frame of a test video.

\subsection{Weak Supervision}

To produce video-level predictions of foreground actions, we perform
attention-weighted average pooling of frame features over the whole video to
produce a single video-level {\em foreground} feature ${\bf x}_{\rm fg}$
given by
\begin{equation}
{\bf x}_{\rm fg}= \frac{1}{T} \sum_{t=1}^{T}\lambda_t{\bf x}_{t}.
\end{equation}
The weighting for each frame is a scalar $\lambda_t \in [0,1]$
which serves to pick out (foreground) frames during which an action is taking
place while down-weighting contribution from background. The attention is a
function of the $d$-dimensional frame feature $\lambda_t=\Omega(x_t)$ which we
implement using two fully-connected (FC) layers with a ReLU activation for the
first layer and a sigmoid activation function for the second.

To produce a video-level prediction, we feed the pooled feature to
fully-connected softmax layer, parameterized by $w_c \in \mathbb{R}^d$ for
class $c$:
\begin{equation}
p_{\rm fg}[c] = \frac{e^{w_c \cdot {\bf x}_{\rm fg}}} {\sum_{i= 0}^C e^{w_i
  \cdot {\bf x}_{\rm fg}}} \label{eq:fg}
\end{equation}
The foreground classification loss is the defined via regular cross-entropy
loss with respect to the video label $y$.
\begin{equation}
\mathcal{L}_{\rm fg}=-\log p_{\rm fg}[y]
\end{equation}

\paragraph{Background-Aware Loss} The complement of the attention factor,
$1-\lambda$, indicates frames where the model believes that no action is taking
place. We propose that features pooled from such background frames ${\bf
x}_{\rm bg}$ should also be classified by the {\em same} softmax model as was
applied to the pooled foreground frames.
\begin{align}
{\bf x}_{\rm bg} &=  \frac{1}{T}\sum_{t=1}^{T}(1-\lambda_{t}){\bf x}_{t}\\
  p_{\rm bg}[c] &= \frac{e^{w_c \cdot {\bf x}_{\rm bg}}} {\sum_{i = 0}^C e^{w_i
  \cdot {\bf x}_{\rm bg}}} \label{eq:bg}
\end{align}
The vector, $p_{\rm bg}\in\mathbb{R}^{C+1}$, indicate the likelihood of each
action class for the background-pooled features. The background-aware loss,
$\mathcal{L}_{\rm bg}$, encourages this vector to be close to $1$ at the
background index, $y=0$, and $0$ otherwise.  This cross entropy loss on the
background feature then simplifies to
\begin{align*}
\mathcal{L}_{\rm bg} = -\log p_{\rm bg}[0]
\end{align*}
Compared to a model which is trained to classify only foreground frames,
$\mathcal{L}_{\rm bg}$,  ensures that the parameters $w$ also learn to
distinguish actions from the background.
\begin{figure*}[ht]
\captionsetup{font=small}
\centering
\includegraphics[width=1\linewidth]{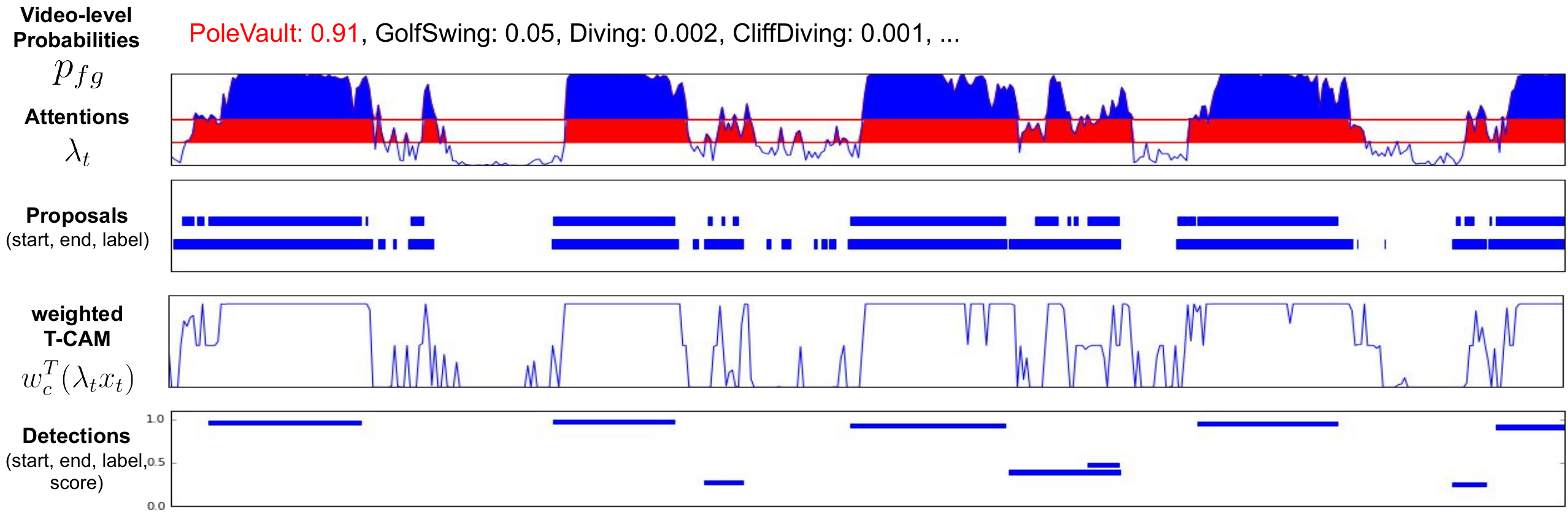}
	\caption{The detection process involves three steps: video-level class
	probability thresholding, segment proposal generation and detection
	scoring. First, relevant classes are selected by thresholding
	video-level probabilities. The attention vector is thresholded with
	different values to select salient, connected segments.  Each threshold
	value corresponds to a different set of segment proposals which are
	pooled.  Each proposal is scored by averaging the weighted-TCAM values
	within its interval. Per-class non-maxima suppression is performed to
	remove highly overlapped detections. The y-axis in last figure
	indicates the final detection score.}
\label{fig:inference_process}
\end{figure*}
\paragraph{Self-guided Attention Loss}
The attention variable $\lambda_t$ can be thought of as a {\em bottom-up}, or
class-agnostic, attention model that estimates the foreground probability of a
frame. This will likely respond to generic cues such as large body motions,
which are not specific to particular actions. Recent works have shown one can
extract {\em top-down} attentional cues from classifiers operating on pooled
features by examining (temporal) class activation maps (TCAM)
~\cite{nguyen2017weakly,zhou16learning}. We propose to use class-specific TCAM
attention maps as a form of self-supervision to refine the class-agnostic
bottom-up attention maps $\lambda_t$. Specifically, we use top-down attention
maps from the class $y$ that is {\em known} to be a given training video:
\begin{align}
\hat{\lambda}^{\rm fg}_t &=G(\sigma)\ast \frac{e^{w_y {\bf x}_t}}{\sum_{i=0}^C
  e^{w_i {\bf x}_t}}
\end{align}
where $G(\sigma)$ refers to a Gaussian filter used to temporally smooth the
class-specific, top-down attention signals.\footnote{If video is labeled with
multiple actions, we max-pool foreground attention targets $\hat{\lambda}^{\rm
fg}_t$ across all present actions so that $\hat{\lambda}_t$ is large if any
action is taking place at time $t$.} Gaussian smoothing imposes the intuitive
prior that if a frame has high probability of being an action, its neighboring
frames should also have high probability of containing an action.  Note the
above softmax differs from \eqref{eq:fg} and \eqref{eq:bg} in that they are
defined at the frame level (as opposed to the video level) and that they are
{\bf not} modulated by bottom-up attention $\lambda_t = \Omega({\bf x}_t)$.

Since our top-down classifier also includes a model of background, we can
consider an attention target given by the complement of the background class
activations
\begin{align}
\hat{\lambda}^{\rm bg}_t &= G(\sigma)\ast \frac{\sum_{i=1}^C e^{w_i {\bf
  x}_t}}{\sum_{i=0}^C e^{w_i {\bf x}_t}}
\end{align}
Given this attention target, we define the self-guided loss as
\[
\mathcal{L}_{\rm guide}=\frac{1}{T}\sum_t|\lambda_t-\hat{\lambda}^{\rm
fg}_t|+|\lambda_{t}-{\hat \lambda}^{\rm bg}_t|
\]
which biases the class-agnostic bottom-up attention map to agree with the
top-down class-specific attention map (for classes known to exist in a given
training video).

\paragraph{Foreground-background Clustering Loss} 
Finally, we consider a bottom-up loss defined purely in terms of the
video features and attention $\lambda$ which makes no reference to the
video-level labels.  We estimate another set of parameters $u_{fg},u_{bg} \in
\mathbb{R}^d$ that are applied to the bottom-up attention-pooled features
(that do not require top-down class labels)
\begin{align}
z_{\rm fg} &= \frac{e^{u_{fg} {\bf x}_{fg}}}{e^{u_{fg} {\bf x}_{fg}} +
  e^{u_{bg} {\bf x}_{fg}}}\\
z_{\rm bg} &= \frac{e^{u_{bg} {\bf x}_{bg}}}{e^{u_{fg} {\bf x}_{bg}} +
  e^{u_{bg} {\bf x}_{bg}}}
\end{align}
Each video should contain both foreground and background frames so the
clustering loss encourages both classifiers respond strongly to their
corresponding pooled features
\begin{equation}
\mathcal{L}_{\rm cluster} = -\log z_{\rm fg} - \log z_{\rm bg}
\end{equation}
This can be viewed as a clustering loss that encourages the foreground and
background pooled features to be distinct from each other.

\paragraph{Total loss}
We combine these losses to yield a total per-video training loss 
\begin{equation}
\mathcal{L}_{total}=\mathcal{L}_{\rm fg}+\alpha \mathcal{L}_{\rm bg}+\beta
\mathcal{L}_{\rm guide}+\gamma \mathcal{L}_{\rm cluster}.
\label{eq:total_loss}
\end{equation}
with $\alpha, \beta$ and $\gamma$ are the hyperparameters to control the
corresponding weights between the losses. We find that these hyperparameters
($\alpha, \beta, \gamma$) need to be small enough so that network is driven
mostly by the foreground loss, $\mathcal{L}_{\rm fg}$.

\subsection{Action Localization}

To generate action proposals and detections, we first identify relevant action
classes based on video-level classification probabilities, $p_{\rm fg}$. Segment
proposals are generated for each relevant class. These proposals are then
scored with the corresponding weighted T-CAMs to obtain the final detections.
\begin{table*}[t]
\captionsetup{font=small}
\centering
\caption{Ablation studies show each additional loss leads to significant
	localization performance gain. The losses also complement each other as
	combining them achieves better results. The first and second rows are
	obtained from STPN~\cite{nguyen2017weakly}.}
\label{table:ablation_studies}
\small
\begin{tabular}{ccccc|ccccccccc}
& & & & & \multicolumn{9}{c}{AP@IoU}  \\
$\mathcal{L}_{fg}$ & $\mathcal{L}_{bg}$ & $\mathcal{L}_{guide}$ &
	$\mathcal{L}_{cluster}$ & $\mathcal{L}_{sparse}$ & 0.1 & 0.2 & 0.3 &
	0.4 & 0.5 & 0.6 & 0.7 & 0.8 & 0.9 \\
\hline
\checkmark & -- & -- & -- & -- & 46.6 & 38.7 & 31.2 & 22.6 & 14.7 & -- & -- & -- & -- \\
\checkmark & -- & -- & -- & \checkmark & 52.0 & 44.7 & 35.5 & 25.8 & 16.9 & 9.9 & 4.3 & 1.2 & 0.2 \\
\checkmark & -- & \checkmark & -- & -- & 53.8 & 46.4 & 38.2 & 29.0 & 19.2 & 10.6 & 4.4 & 1.3 & 0.1 \\
\checkmark & \checkmark & -- & -- & -- & 53.6 & 47.6 & 39.1 & 30.2 & 20.5 & 12.2 & 5.4 & 1.7 & 0.2 \\
\hline
\checkmark & \checkmark & \checkmark & -- & -- & 58.9 & 54.3 & 41.5 & 33.9 & 24.4 & 16.2 & 7.8 & 2.4 & 0.4 \\
\checkmark & \checkmark & -- & \checkmark & -- & 54.9 & 48.4 & 40.8 & 32.4 & 23.1 & 14.2 & 7.4 & 2.5 & 0.3 \\
\checkmark & -- & \checkmark & \checkmark & -- & 60.1 & 54.1 & 45.6 & 34.0 & 23.2 & 13.6 & 6.2 & 1.4 & 0.1 \\
\hline
\checkmark & \checkmark & \checkmark & \checkmark & -- & 60.4 & 56.0 & 46.6 & 37.5 & 26.8 & 17.6 & 9.0 & 3.3 & 0.4 \\
\hline
\end{tabular}
\end{table*}
We keep segment-level features at timestamp $t$ with attention value
$\lambda_t$ greater than some pre-determined threshold. We perform 1-D
connected components for connect neighboring segments to form segment proposal.
A segment proposal $[t_{start},t_{end},c]$, is then scored as
\begin{equation}
\sum_{t=t_{\text{start}}}^{t_{\text{end}}}\frac{\theta\lambda^{RGB}_{t}w_{c}^{T}{\bf
  x}_{t}^{RGB}+(1-\theta)\lambda_t^{FLOW}w_{c}^{T}
{\bf x_t} ^{FLOW}}{t_{\text{end}}-t_{\text{start}}+1}
\end{equation}
\noindent  where $\theta$ is a scalar denoting the relative importance between
the modalities. In this work, we set $\theta=0.5$.

Figure~\ref{fig:inference_process} shows an example of the inference process.
Unlike STPN, we do not generate proposals using attention-weighted T-CAMs but
from the attention vector, ${\bf \lambda}$. Multiple thresholds are used to
provide a larger pool of proposals. We find that generating proposals from the
averaged attention weights from different modalities leads to more reliable
proposals. Class-wise non-maxima suppression (NMS) is used to remove detections
with high overlap.

\section{Experiments}
\subsection{Datasets and Evaluation Method}
\label{sub:datasets}
\noindent {\bf Datasets} We evaluate the proposed algorithm on two popular
action detection benchmarks, THUMOS14~\cite{jiang14thumos} and
ActivityNet1.3~\cite{activitynet}.

THUMOS14 has temporal boundary annotations for 20 action classes in 212 validation
videos and 200  test videos. Following standard protocols,
we train using the validation subset without temporal annotations and evaluate
using test videos. Video length ranges from a few
seconds up to 26 minutes, with the mean duration around 3 minutes long. On
average, there are 15 action instances per video.  There is also a large
variance in the length of an action instance, from less than a second to
minutes.
\begin{figure*}[ht]
\captionsetup{font=small}
\centering
\includegraphics[width=1\linewidth]{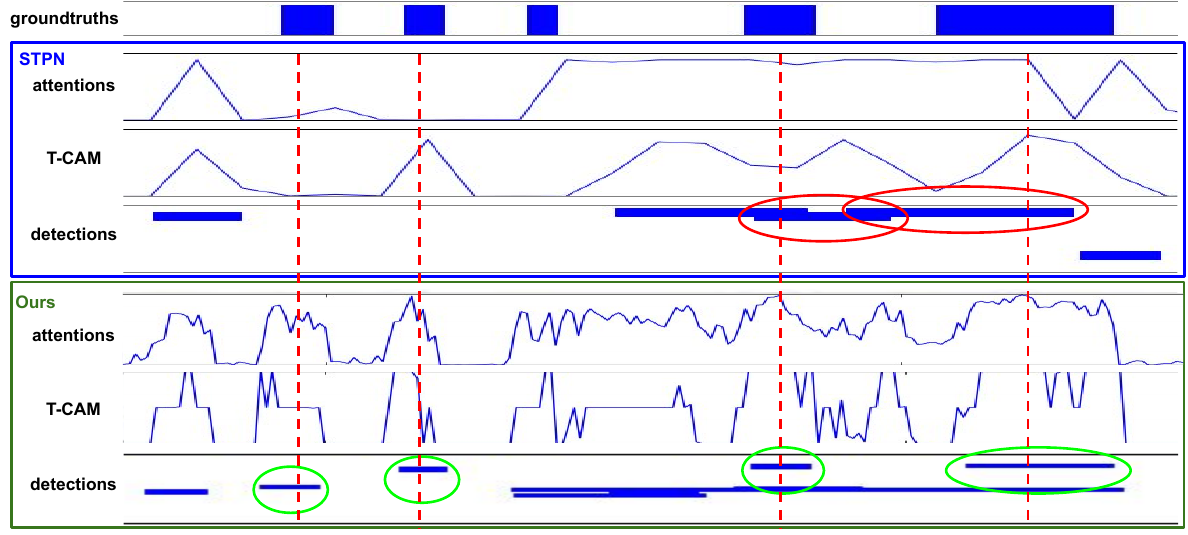}
    \caption{With background modeling, our model is able to produce better
	attention weights, T-CAM signals and subsequently better detections.
	The first two action instances (green ellipses) are detected by our
	methods but completely missed by STPN. While both algorithms detect the
	last two action instances (last red and green ellipses), ours is able
	to obtain more accurate boundaries.}
    \label{fig:comparisons_with_STPN}
\end{figure*}

The ActivityNet dataset offers a larger benchmark for complex action
localization in untrimmed videos. We use ActivityNet1.3, which has 10,024
videos for training, 4,926 for validation, and 5,044 for testing with 200
activity classes. For fair comparisons, we use the same pre-extracted I3D
features as STPN.

Microvideos are short, untrimmed video clips available on social media
platforms, such as Instagram and Snapchat. These videos are authored to
be exciting, and hence often have much higher foreground/background content
ratio than regular videos. We aim to leverage this new source of data and its
accompanying tags to improve action localization performance. We download $100$
most-recent Instagram videos containing tags constructed from THUMOS14's action
names. For example, for {\it `BaseballPitch'}, we query Instagram for videos
with tag {\it \#baseballpitch}. Duplicated and mis-tagged videos are removed.
The retention rate depends on the action labels, ranging from $15\%$ to $89\%$
with the average retention rate at $45\%$.  It takes less than $2$ hours to
curate video-level labels for $2000$ videos.  The final set contains a total of
$915$ microvideos.  The duration for these videos ranges from $6$ to $15$
seconds. Each video often 1-2 action instances. Example microvideos
are shown in our supplementary materials. In our experiments, we simply add these
microvideos to the THUMOS14 train set and keep the rest of the experiment
unchanged.

We follow the standard evaluation protocol based on mean Average Precision
(mAP) values at different levels of intersection over union (IoU) thresholds.
The evaluation is conducted using the benchmarking code for the temporal action
localization task provided by
ActivityNet\footnote{https://github.com/activitynet/ActivityNet/blob/master/Evaluation/}.

\subsection{Implementation Details} For fair comparisons, experiment
settings are kept similar to STPN~\cite{nguyen2017weakly}. Specifically, we use
two-stream I3D networks trained on Kinetics~\cite{kay2017kinetics} as
segment-level feature extractor.  I3D features are extracted using
publicly-available code and
models\footnote{https://github.com/deepmind/kinetics-i3d}. We follow the
preprocessing steps for RGB and optical flow recommended by the software. For
the flow stream, we use an OpenCV implementation to calculate the dense optical
flow using the Gunnar Farneback's algorithm~\cite{farneback2003two}. Instead of
sampling a fixed number of segments per video like STPN, we load all the
segments for one video and process only one video per batch.

The loss function weights in Eq.~\ref{eq:total_loss} are set as
$\alpha=\beta=\gamma=0.1$. This specific setting is provided for ease of
reproducibility. However, as long as these values are around 10x smaller than
foreground class loss weight, converged models have similar performance.
Intuitively, video-level labels provide the most valuable supervision. The
higher foreground class loss weight encourages the model to first produce
correct video-level labels. Once the foreground loss is saturated, minimizing
the other losses improves boundary decisions between foreground and background.

The network is implemented in TensorFlow and trained using the Adam optimizer
with learning rate $10^{-4}$.  At testing time, we reject classes whose
video-level probabilities are below $0.1$. If no foreground class has
probability great than $0.1$, we generate proposals and detections for the
highest foreground class. We propose using a large set of thresholds ranging
from $0$ to $0.5$ with the $0.025$ increment. All proposals are combined in one
large set. We use an NMS overlap threshold of $0.5$. 


\section{Results}
\begin{table*}[t]
\captionsetup{font=small}
\centering
  \caption{Comparisons with recent techniques on THUMOS14. Our method yields 
  ~10\% improvement over the original system~\cite{nguyen2017weakly}. We
  significantly outperform other weakly supervised
  approaches~\cite{shou2018autoloc,paul2018w}, $5\%$ mAP@$0.5$. In general, our
  model performance is comparable to fully-supervised methods in lower IoU regimes. Higher
  IoU requires more accurate action boundary decisions, which is difficult to
  do without the actual boundary supervision.}
\vspace{-0.2cm}
\label{table:thumos14_results}
\small
\begin{tabular}{c|c||ccccccccc}
\multirow{2}{*}{Supervision} & \multirow{2}{*}{Method} & \multicolumn{9}{c}{AP@IoU}  \\
&  & 0.1 & 0.2 & 0.3 & 0.4 & 0.5 & 0.6 & 0.7 & 0.8 & 0.9 \\
\hline
\multirow{11}{*}{\shortstack{Fully \\ supervised}} & Heilbron et al.~\cite{heilbron16fast} & -- & -- & -- & -- & 13.5  & -- & -- & -- & -- \\
& Richard et al.~\cite{richard16temporal}  & 39.7 & 35.7 & 30.0 & 23.2 & 15.2 & --& -- & -- & -- \\
& Shou et al.~\cite{shou16temporal}  	& 47.7 & 43.5 & 36.3 & 28.7 & 19.0 & 10.3 & {\color{white}0}5.3 & -- & -- \\
& Yeung et al.~\cite{yeung16end}  	& 48.9 & 44.0 & 36.0 & 26.4 & 17.1 & -- & -- & -- & -- \\
& Yuan et al.~\cite{yuan16temporal}  	& 51.4 & 42.6 & 33.6 & 26.1 & 18.8 & -- & -- & -- & -- \\
& Escordia et al.~\cite{escorcia16daps}	& -- & -- & -- & -- & 13.9 & -- & -- & -- & -- \\
& Shou et al.~\cite{shou17cdc}		& -- & -- & 40.1 & 29.4 & 23.3 & 13.1 & {\color{white}0}7.9 & -- & -- \\
& Yuan et al.\cite{yuan17temporal}	& 51.0 & 45.2 & 36.5 & 27.8 & 17.8 & -- & -- & -- & -- \\
& Xu et al.\cite{xu17r}		& 54.5 & 51.5 & 44.8 & 35.6 & 28.9 & -- & -- & -- & -- \\
& Zhao et al.~\cite{zhao2017temporal} & 66.0 & 59.4 & 51.9 & 41.0 & 29.8 & -- & -- & -- & -- \\
& Chao et al.~\cite{chao2018rethinking} & 59.8 & 57.1 & 53.2 & 48.5 & 42.8 & 33.8 & 20.8 & -- & -- \\
& Alwassel et al.~\cite{alwassel2018action} & -- & -- & 51.8 & 42.4 & 30.8 & 20.2 & 11.1 & -- & --\\
\hline
\multirow{3}{*}{\shortstack{Weakly \\ supervised}} & Wang~\etal~\cite{wang17untrimmednets} & 44.4 & 37.7 & 28.2 & 21.1 & 13.7 & -- & -- & -- & -- \\
& Singh \& Lee~\cite{singh17hide} & 36.4 & 27.8 & 19.5 & 12.7 & {\color{white}0}6.8 & -- & -- & -- & -- \\
& Nguyen~\etal.\cite{nguyen2017weakly} & 52.0 & 44.7 & 35.5 & 25.8 & 16.9 & 9.9 & 4.3 & 1.2 & 0.2 \\
& Paul~\etal.\cite{paul2018w} & 55.2 & 49.6 & 40.1 & 31.1 & 22.8 & -- & 7.6 & -- & -- \\
& Shou et al.\cite{shou2018autoloc} & -- & -- & 35.8 & 29.0 & 21.2 & 13.4 & 5.8 & -- & -- \\
& Ours 	   & 60.4 & 56.0 & 46.6 & 37.5 & 26.8 & 17.6 & 9.0 & 3.3 & 0.4 \\
& Ours + MV  & 64.2 & 59.5 & 49.1 & 38.4 & 27.5 & 17.3 & 8.6 & 3.2 & 0.5 \\
\hline
\end{tabular}
\end{table*}
\begin{table}[t]
\captionsetup{font=small}
\centering
  \caption{Results on the ActivityNet1.3 validation set.}
\label{table:comparison_activitynet_validation}
\vspace{-0.2cm}
\small
\scalebox{0.9}{
\begin{tabular}{c|c||ccc}
\multirow{2}{*}{} & \multirow{2}{*}{Method} & \multicolumn{3}{c}{AP@IoU}  \\
& & 0.5 & 0.75 & 0.95 \\
\hline
\multirow{7}{*}{\shortstack{Fully \\ supervised}} & Singh \&
  Cuzzolin~\cite{singh16untrimmed}& 34.5 & -- 	& -- \\
  & Wang \& Tao~\cite{wang16anet} 	& 45.1 & 4.1 	& 0.0 \\
  & Shou~\etal~\cite{shou17cdc} 	& 45.3 & 26.0 	& 0.2 \\
  & Xiong~\etal~\cite{zhao2017temporal} 	& 39.1 & 23.5 	& 5.5 \\
  \cline{2-5}
  & Montes~\etal~\cite{montes16temporal}	& 22.5 & -- 	& -- \\
  & Xu~\etal~\cite{xu2017r}	& 26.8 & -- 	& -- \\
	& Chao~\etal~\cite{chao2018rethinking}	& 38.2 & 18.3 	& 1.30 \\
\hline
\multirow{2}{*}{Weakly supervised}	& Nguyen et al.~\cite{nguyen2017weakly}
	& 29.3	& 16.9 	& 2.6\\
	& Ours & 36.4	& 19.2 	& 2.9\\
\hline
\end{tabular}
}
\end{table}
\noindent We perform ablation studies on different combinations of loss
terms to further understand the contribution of each loss. Results in
Table~\ref{table:ablation_studies} suggest the addition of each loss improves
localization performance. Combining these losses in training leads to even
better results, implying that each provides complementary cues.

Figure~\ref{fig:comparisons_with_STPN} shows an example comparing the
intermediate outputs between our model and STPN.  Our model is able to produce
better attentions, T-CAMs, and, consequently, better action detections. Our
model is able to detect instances that are completely missed by the previous model.
This leads to an overall improvement in the recall rate and average precision
of the localization model across different IoU overlap thresholds. For action
instances detected by both models, our model is able to obtain more accurate
temporal boundaries. This leads to AP improvements for stricter IoU overlap thresholds.

\noindent {\bf Comparisons with state-of-the-art}
Table~\ref{table:thumos14_results} compares the action localization results
of our approach on THUMOS14 to other weakly-supervised and fully-supervised
localization systems published in the last three years. For IoUs less than $0.5$,
we improve mAP by ~$10\%$ mAP over STPN~\cite{nguyen2017weakly}. We also
significantly outperform more recent state-of-the-art weakly-supervised action
localization systems. Our model is also comparable to other fully-supervised
systems, especially in the lower IoU regime. In higher IoU overlap regimes, our
model doesn't perform as well as Chao~\etal~\cite{chao2018rethinking}. This
suggests that our model knows where actions happen, but is not able to
precisely articulate the boundaries as well as fully-supervised methods. This
is reasonable as our weakly-supervised models are not privy to boundary
annotations for which fully-supervised methods have full access.

Table~\ref{table:comparison_activitynet_validation} compares our results
against other state-of-the-art approaches on the ActivityNet 1.3 validation
set. Similar to THUMOS14, our method significantly outperforms existing
weakly-supervised approaches while maintaining competitive with other
fully-supervised methods.
\begin{figure*}[t]
\captionsetup{font=small}
\centering
    \begin{subfigure}[b]{\textwidth}
    	\includegraphics[width=\textwidth]{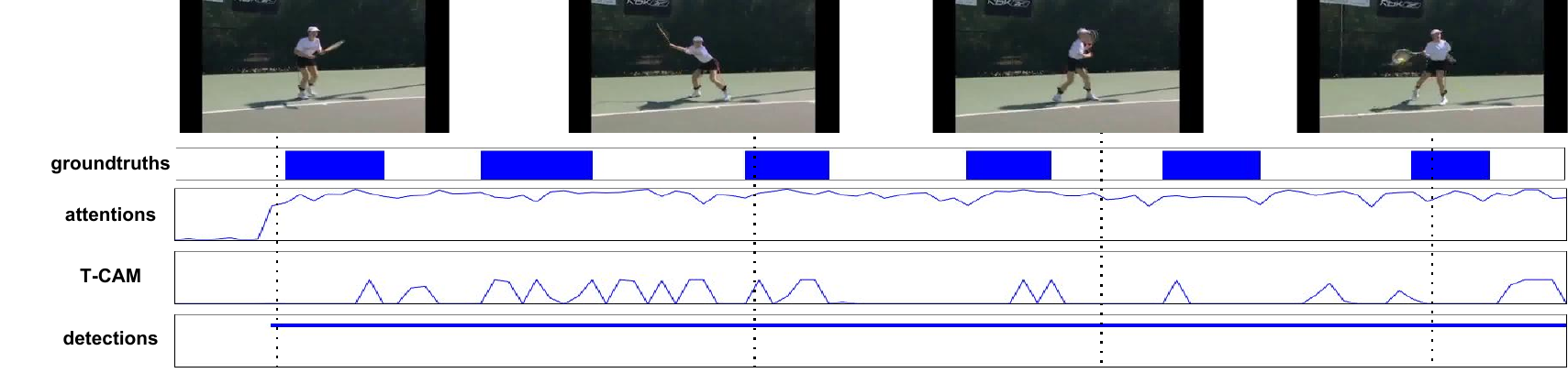}
      \caption{Failure due to similar background across consecutive instances
	    ({\it Tennis}).}
    	\label{fig:failure_cases_difficult_background}
    \end{subfigure}
    \begin{subfigure}[b]{\textwidth}
      \includegraphics[width=\textwidth]{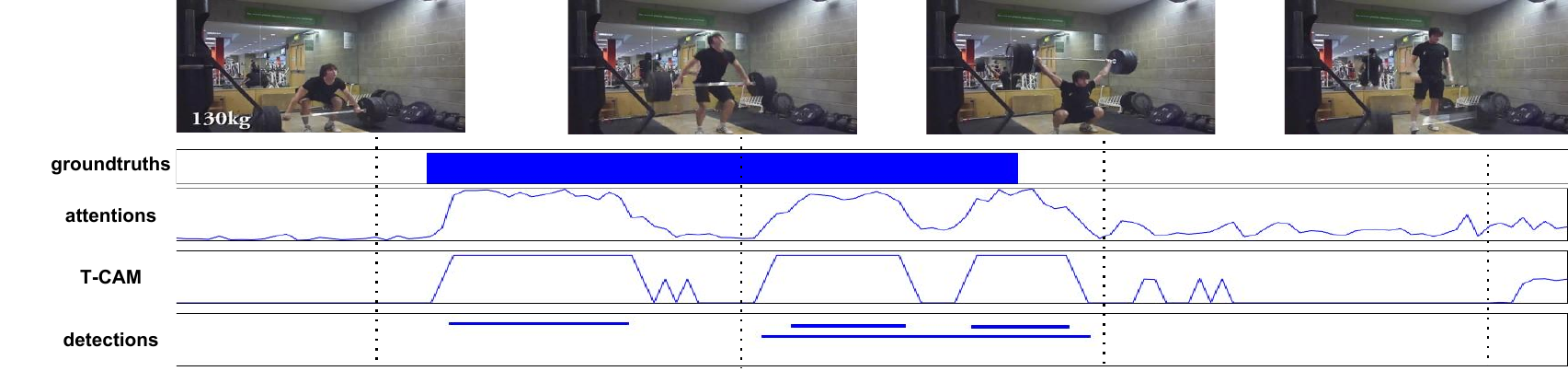}
      \caption{Failure due to action composed of two mini-actions ({\it
	    CleanAndJerk}).}
      \label{fig:failure_cases_action_composition}
    \end{subfigure}
    \begin{subfigure}[b]{\textwidth}
      \includegraphics[width=\textwidth]{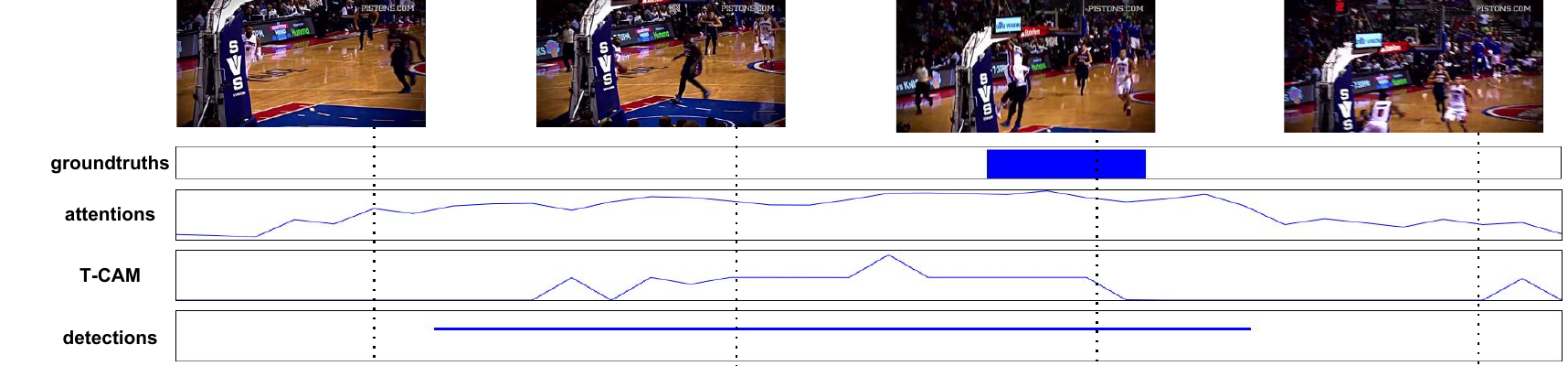}
      \caption{Failure due to subjective boundaries ({\it Basketball}).}
      \label{fig:failure_cases_subjective_boundaries}
    \end{subfigure}
    \caption{Qualitative examples of failure cases where it is
    difficult to resolve action locations with only video-level supervision.}
\label{fig:failure_cases}
\vspace{-0.6cm}
\end{figure*}

\noindent {\bf Micro-videos as supplemental training data} Even though THUMOS14
has a uniform number of training videos across each action class, the class
distribution of action instances is heavily skewed (ranging from 30 instances
of {\it BaseballPitch} to 499 instances of {\it Diving}). As a result,
categories with higher instance count ({\it Diving}, {\it HammerThrow}) have
higher mAP while those with fewer action instances ({\it BaseballPitch}, {\it
TennisSwing}, {\it CleanAndJerk}) have lower mAP. The addition of microvideos
re-balances the skewed class distribution for action instances and improves the
generalizability for categories with lower action instance count.  We observe
improvements of at least $3\%$AP@IoU=$0.5$ for $5$ action categories with the
lowest instance count.

Table~\ref{table:thumos14_results} shows models trained with additional
microvideos (`Ours + MV') improve significantly for IoU thresholds from $0.1$
to $0.5$, while maintaining similar performance at the higher IoU regime.  This
suggests the addition of microvideos allows models to recognize action
instances better, but does not help with generating highly precise boundaries.
These results, along with the ease of collecting and curating  microvideos,
presents a promising direction of using microvideos as a weakly-supervised
training supplement for actional localization.

\noindent {\bf Failure modes} Figure~\ref{fig:failure_cases} examines current
failure modes of our approach.
Figure~\ref{fig:failure_cases_difficult_background} shows multiple action
instances happening close to each other, with little or no background between
them.  When little background happens between actions, the model fails to
correctly split the actions.  Figure~\ref{fig:failure_cases_action_composition}
shows an example of composite actions {\it CleanAndJerk}. The person performing
these action usually stands still between these actions, hence the model breaks
this into two components.  In
Figure~\ref{fig:failure_cases_subjective_boundaries}, we see another
difficulty, namely the subjectivity of boundary annotations. In training
videos, the action of `BasketballDunk' usually involves someone running to the
basket, jumping and dunking the ball. Human annotations however just consider
the last piece of the action as the ground-truth. It is challenging for
weakly-supervised methods to find the correct human-agreed boundaries in this
case, limiting performance in higher IoU regimes. For a better visual sense of
these failure cases, we refer the reader to our supplementary materials.

{\bf Discussion} Without sparsity loss, the majority of STPN's attention weights
$\lambda_t$ remain close to $1$, rendering them useless for detection generation.
The sparsity loss forces the attention module to output more diverse values for
attention weights. However, this loss in combination with video-level
foreground loss encourages the model to select the smallest number of frames
necessary to predict the video-level labels. After a certain point in the
training process, localization performance starts to deteriorate significantly
as the sparsity loss continues to eliminate relevant frames.  {\em This requires
early stopping to prevent performance drop}. In contrast, our model uses
top-down T-CAMs as a form of self-supervision for the attention weights. As a
result, our model can simply be trained to convergence.

{\bf Conclusion} We introduced a method for learning action localization from
weakly supervised training data which outperforms existing approaches and even
some fully supervised models. We attribute the success of this approach to
building up an explicit model for background content in the video. By coupling
top-down models for action with bottom-up models for clustering, we are able to
learn a latent attention signal that can be used to propose action intervals
using simple thresholding without the need for more complex sparsity or
temporal priors on action extent. Perhaps most exciting is that the resulting
model can make use of additional weakly supervised data which is readily
collected online. Despite domain shift between Instagram videos and THUMOS14,
we are still able to improve performance across many categories, demonstrating
the power of the weakly supervised approach to overcome the costs associated
with expensive video annotation.

{\bf Acknowledgements} This work was supported in part by a hardware donation
from NVIDIA,  NSF Grants 1253538, 1618903 and the Office of the Director of
National Intelligence (ODNI), Intelligence Advanced Research Projects Activity
(IARPA),  via Department of Interior/Interior Business Center (DOI/IBC)
contract number D17PC00345. The U.S. Government is authorized to reproduce and
distribute reprints for Governmental purposes not withstanding any copyright
annotation theron. Disclaimer: The views and conclusions contained herein are
those of the authors and should not be interpreted as necessarily representing
the official policies or endorsements, either expressed or implied of IARPA,
DOI/IBC or the U.S. Government.

{\small
\bibliographystyle{ieee_fullname}
\bibliography{egbib}
}

\end{document}